\documentclass[letterpaper, 10 pt, conference]{ieeeconf}  % Comment this line out if you need a4paper

\IEEEoverridecommandlockouts                              % This command is only needed if 
\usepackage[
backend=bibtex,
style=numeric-comp,
sorting=none
]{biblatex}
\usepackage{graphicx} % for pdf, bitmapped graphics files
\usepackage[table]{xcolor}% http://ctan.org/pkg/xcolor
\usepackage{color}
\usepackage{tabularx}
\usepackage{booktabs}
\usepackage{multirow}
\usepackage{subcaption}
\usepackage{xspace}

\newcommand{\etal}{et al.\@\xspace}

\title{\LARGE \bf
SHOPPER: Practical Insights on Grasp Strategies for \\Mobile Manipulation in the Wild}

\author{Isabella Huang*, Richard Cheng*, Sangwoon Kim*, Dan Kruse*, Carolyn Chen*, \\ Lukas Kaul, JC Hancock, Shanmuga Harikumar, Mark Tjersland, James Borders, Dan Helmick \\  \normalsize Toyota Research Institute, Los Altos, California
\thanks{$^{*}$These authors contributed equally to this work}%
}

\begin{document}

\maketitle
\thispagestyle{empty}
\pagestyle{empty}

%%%%%%%%%%%%%%%%%%%%%%%%%%%%%%%%%%%%%%%%%%%%%%%%%%%%%%%%%%%%%%%%%%%%%%%%%%%%%%%%
\begin{abstract}

Mobile manipulation systems have advanced significantly in recent years. However, substantial gaps remain that prevent state-of-the-art platforms from achieving widespread real-world deployment, particularly in reliably grasping items in unstructured environments. To help bridge this gap, we develop SHOPPER, a mobile manipulation robot platform designed to push the boundaries of reliable and generalizable grasp strategies. We develop these grasp strategies and deploy them in a real-world grocery store -- an exceptionally challenging setting chosen for its vast diversity of manipulable items, fixtures, and layouts. In this work, we present our detailed approach to designing general grasp strategies towards picking \textit{any} item in a real grocery store. Additionally, we provide an in-depth analysis of our latest real-world field test, discussing key findings related to fundamental failure modes over hundreds of distinct pick attempts. Through our detailed analysis, we aim to offer valuable practical insights and identify key grasping challenges, which can guide the robotics community towards pressing open problems in the field. Lastly, we provide a dataset of 1200+ grasp attempts in unseen grocery stores.

\end{abstract}

\section{INTRODUCTION}

Grasping and extracting diverse, novel objects from densely cluttered environments is a fundamental challenge in mobile manipulation and a key requirement for real-world robot deployment. Many impressive recent works have shown mobile manipulators grasping a diversity of items in lab settings. However, many grasping works abstract away different parts of the robot stack, leading to assumptions that do not hold in the real-world (e.g. perfect perception, known obstacle geometries, etc.), and exhibit very clean environments or ``structured'' clutter that is not representative of the realistic clutter. As a result, few works have (1) been able to make the jump to the real world or (2) exhibited reliability close to necessary for real-world
deployment. This is reflected in the dearth in widespread deployments of commercial mobile manipulators.  

Because of the varying assumptions made for different systems, one of the major challenges facing the field is even \textit{identifying} the difficult problems and primary bottlenecks for robot grasping in the wild. Furthermore, the lack of large-scale testing naturally creates an increased focus on complexity, while hindering focus on reliability (arguably one of the most important metrics for deployment). Therefore, one of the main contributions of this paper is to address the key question: \textit{What are critical practical considerations and current bottlenecks for improving robot grasps and extracts for mobile manipulators?}

\begin{figure}
\centering
\includegraphics[width=0.48\textwidth]{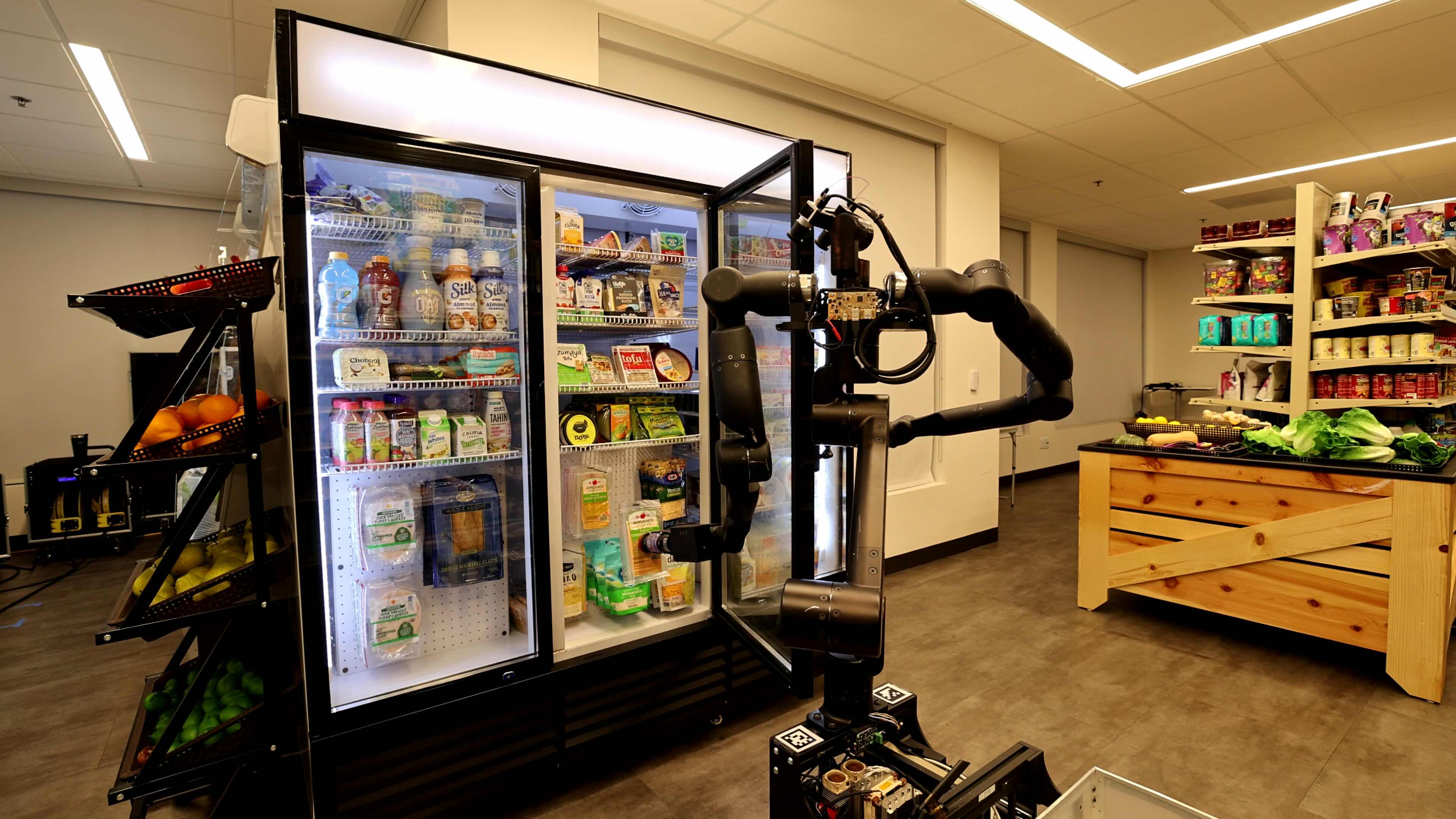}
\caption{SHOPPER is a general-purpose mobile manipulator capable of executing complex grasp strategies, such as retrieving an item hanging on a hook behind a fridge door.}
\label{fig:front_fig}
\end{figure}

In this paper, we present the grasping pipeline for SHOPPER, an autonomous mobile manipulation robot designed for in-the-wild grocery store shopping. We do not abstract away any part of the robot stack (i.e. we run perception, motion planning, navigation, etc. all on-robot and online), and we operate in an \textit{unmodified} real-world grocery store. While the non-grasping portions of the stack are outside the scope of this paper (see \cite{MMT2023} and Fig. \ref{fig:system_diagram}), the fact that they run online ensures that we are dealing with realistic in-the-wild conditions for the robot. This fully integrated system serves as a state-of-the-art research platform for continuous testing and development, and the focus on the grocery store allows us to easily measure quantitative success metrics, enabling practical insights and data-driven development. 

We then conduct a deep-dive into the intricacies and failure modes of grasping from large-scale multi-day testing in a real-world grocery store consisting of thousands of unique items. We analyze hundreds of distinct pick attempts, discussing the critical elements for success and important factors contributing to failures. The primary contributions of this paper are:
\begin{itemize}
    \item A description of our grasping pipeline,
    \item Extensive failure analysis, elucidating practical considerations that are key for successful grasping,
    \item Identification of existing bottlenecks preventing reliable grasping for mobile manipulation in the wild.
    \item A dataset of 1200+ grasp attempts over 800+ unique items from testing in unseen grocery stores\footnote{\url{https://tri-ml.github.io/shopper}}.
\end{itemize}

\section{RELATED WORK}

Robotic mobile manipulation is a dynamic and multifaceted research field, driven by its wide range of potential applications. Interest in this area has been further fueled by international competitions \cite{Sun2022RAL} such as DARPA’s robotics challenge \cite{Guizzo-2015-Spectrum}, RoboCup@Home \cite{wisspeintner-2009-robocup}, the Amazon Picking Challenge \cite{eppner2016lessons} and RoboCup@Work \cite{Kraetzschmar-2015-Robo}, each focusing on different challenges and performance criteria.

Numerous advanced mobile manipulation platforms have emerged from past and ongoing research efforts. Notable examples of wheeled dual-arm mobile manipulators include Willow Garage's PR2 \cite{Meeussen-2010-ICRA}, CMU's Herb2.0 \cite{herb2_journal}, DLR's Rollin' Justin \cite{borst-2009-rollin}, JPL's RoboSimian \cite{robosimian}, KIT's ARMAR-6 \cite{asfour-2019-armar} and IIT's CENTAURO \cite{centauro}.
Comprehensive overviews of wheeled mobile manipulation systems and their associated challenges are provided in \cite{Thakar-2023-MaR} and \cite{Sereinig-2020}.
Most relevant to our research are studies that evaluate mobile manipulation systems in semi-structured environments beyond the lab. Spahn \etal \cite{Spahn2024RSS} also explore item retrieval in a supermarket setting. However, their approach is limited to handling items from open shelving.  D{\"o}mel \etal \cite{domel-2017-toward} used a wheeled, single arm mobile manipulation system to fulfill fetch and carry tasks in a factory environment, similar to our shopping scenario, conducting a full-day evaluation to assess system performance. Compared to prior works, our system features more advanced hardware capabilities and enables a wider range of manipulation tasks, accommodating a greater variety of environmental constraints and obstacles. We also note that other large-scale grasping-focused studies typically do not involve mobile manipulators, and also often assume simplified tabletop environments \cite{Fang2020CVPR, Cao2021RAL}.

End-to-end learning techniques for mobile manipulation are also rising in popularity. Recent works include learning to manipulated articulated objects through behavior cloning \cite{xiong2024adaptivemobilemanipulationarticulated}, and learning to jointly optimize navigation and manipulation for hybrid tasks such as door opening and table wiping \cite{Yang2024IROS}. While modular systems like SHOPPER are typically more robust and generalizable than end-to-end systems currently are, similar large-scale studies as the one presented in this work should also be conducted for end-to-end policies in the future. 

\section{SYSTEM DESIGN}
% We provide a brief overview of the hardware and software design elements of our stack. 
% Further details on all subsystems are available in prior work \cite{MMT2023}. 
\subsection{Hardware}
Our mobile manipulator robot comprises four distinct parts:  (1) a pseudo-holonomic 4-wheeled chassis; (2) a 5-DoF torso; (3) two 7-DoF arms; and (4) a camera pair on a pan-tilt neck. To maximize pick diversity, the right arm is equipped with a Robotiq 2F-85 gripper, while the left arm features a custom suction tool that adjusts vacuum level and flow rate based on internal pressure \cite{MMT2023}. This tool extends suction capabilities to heavier and more deformable objects, handling shear loads up to $\approx$2 kg and vertical lifts up to $\approx$6 kg. For visual perception, we employ two pairs of Basler acA2500-60uc color cameras with wide-angle lenses—one mounted on the pan-tilt neck and another on the chassis front. Each arm is also fitted with a Sunrise Instruments M3553E 6-axis force/torque sensor at the wrist. All computations are managed by a central system featuring an Intel Core i9-12900K CPU and an NVIDIA A6000 GPU.

\subsection{Software}
\begin{figure}
\centering
\includegraphics[scale=0.4]{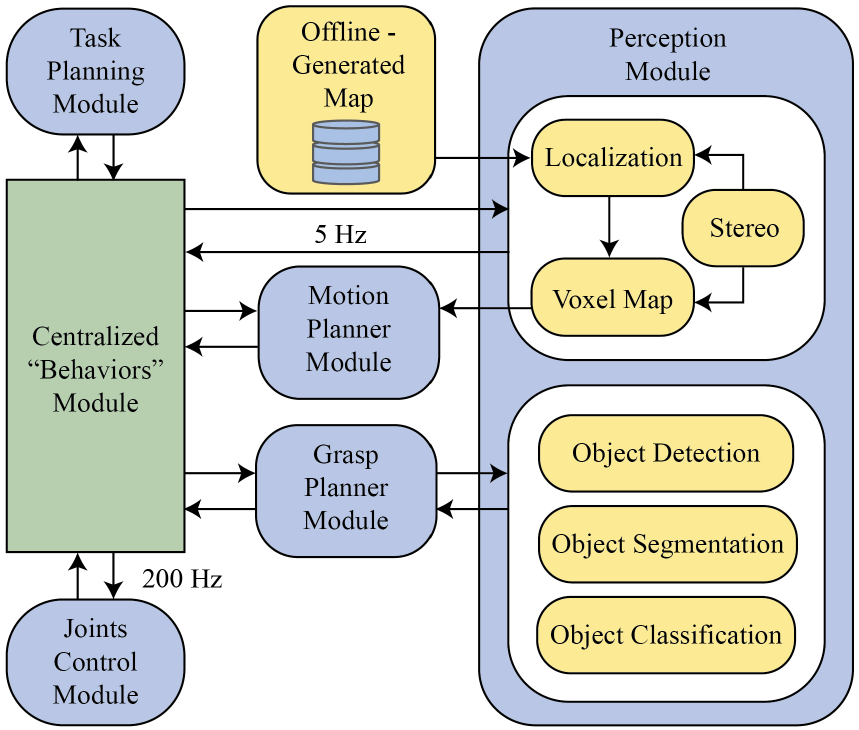}
\caption{Overview of our modular, interconnected software system for mobile manipulation.} 
\label{fig:system_diagram}
\end{figure}

Figure~\ref{fig:system_diagram} shows an overview of our software stack, with different modules running at varying rates, all communicating through a custom inter-process communication (IPC) framework. At the highest level, the task planning module operates a hierarchical finite state machine to determine the robot's next high-level action (e.g., navigate to item, detect item, plan grasp). A centralized behavior module queries relevant sensing and planning modules and sends low-level actuator commands at 200 Hz. The perception module runs at 5 Hz, generating an updated stereo point cloud and voxel map of the surroundings, while the localization module, also at 5 Hz, localizes the robot within an offline-generated map.

\section{TASK}
The task for our mobile manipulation robot is to autonomously retrieve a random shopping list of 20 unique items (out of $\approx$723 mapped items in the grocery store). We also randomize whether to pick 1 or 2 instances of each item. We evaluate the robot's performance in a ``field test'' in a real grocery store, in which the robot continuously runs the task described above, with the shopping list randomized after every run. Each field test spans multiple nights with approximately 10 hours of robot runtime in total.

We deliberately chose our robot's task to focus on grocery stores for several reasons.
\begin{itemize}
    \item Grocery stores feature a huge variety of items, more so than most other environments.
    \item They feature unique real-world challenges (e.g., item density, narrow navigable areas, tight pick points) that are characteristic of settings designed for humans rather than for the ease of robot manipulation.
    \item We can easily gather quantitative metrics to inform our development and measure our performance.
\end{itemize}
We operate in the grocery store after-hours, but we do not alter the environment in any way to ensure that we can operate in a challenging real-world environment.

\subsection{Building the offline map}
\begin{figure}
    \centering
    \begin{subfigure}{0.22\linewidth}
        \centering
        \includegraphics[width=20mm,scale=0.05, trim={0 0 0 0},clip]
        {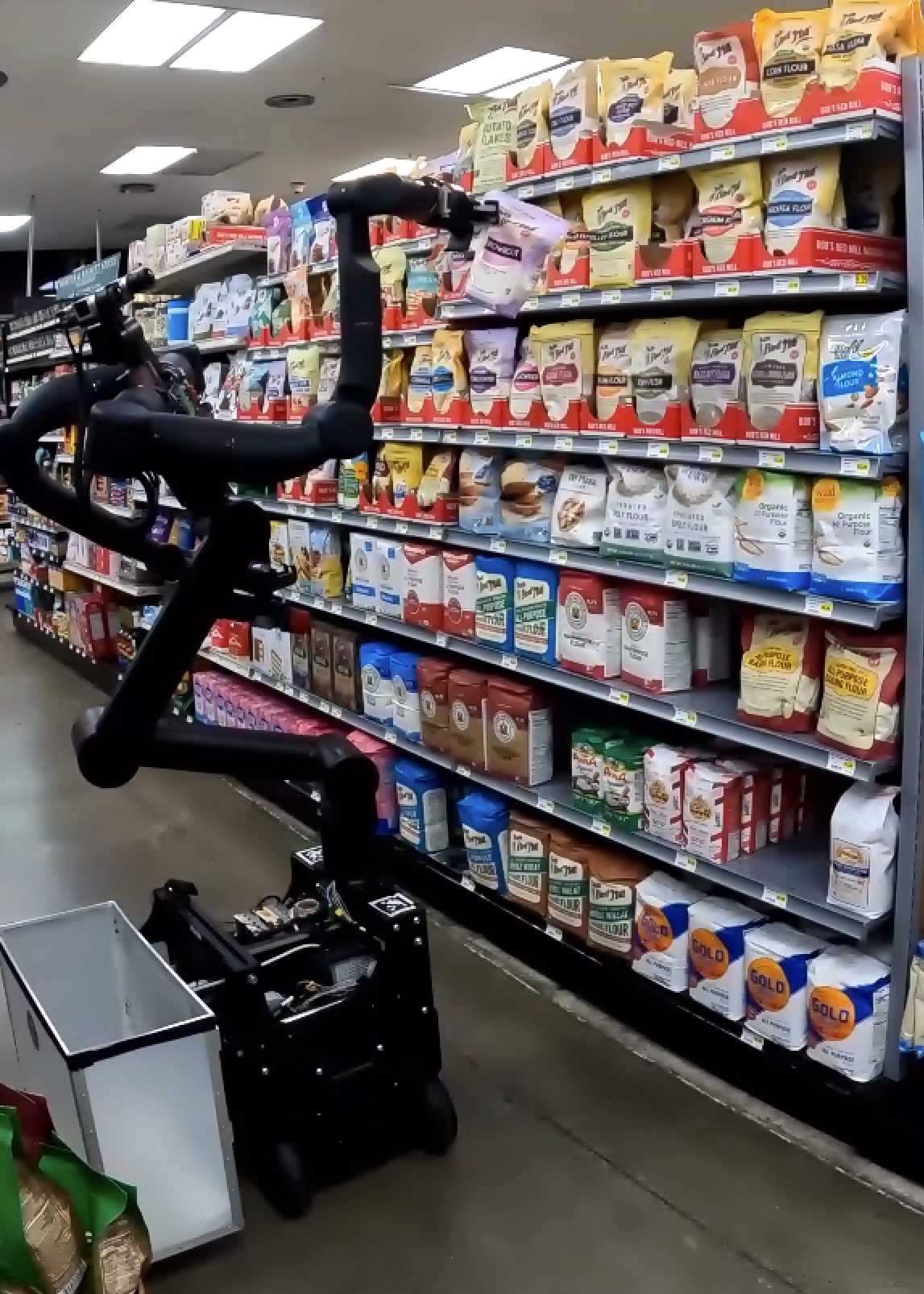}
        \caption{Shelves} 
        \label{fig:layout_a}
    \end{subfigure}%
    ~ 
    \begin{subfigure}{0.22\linewidth}
        \centering
        \includegraphics[width=20mm,scale=0.05, trim={0 0 0 0},clip]{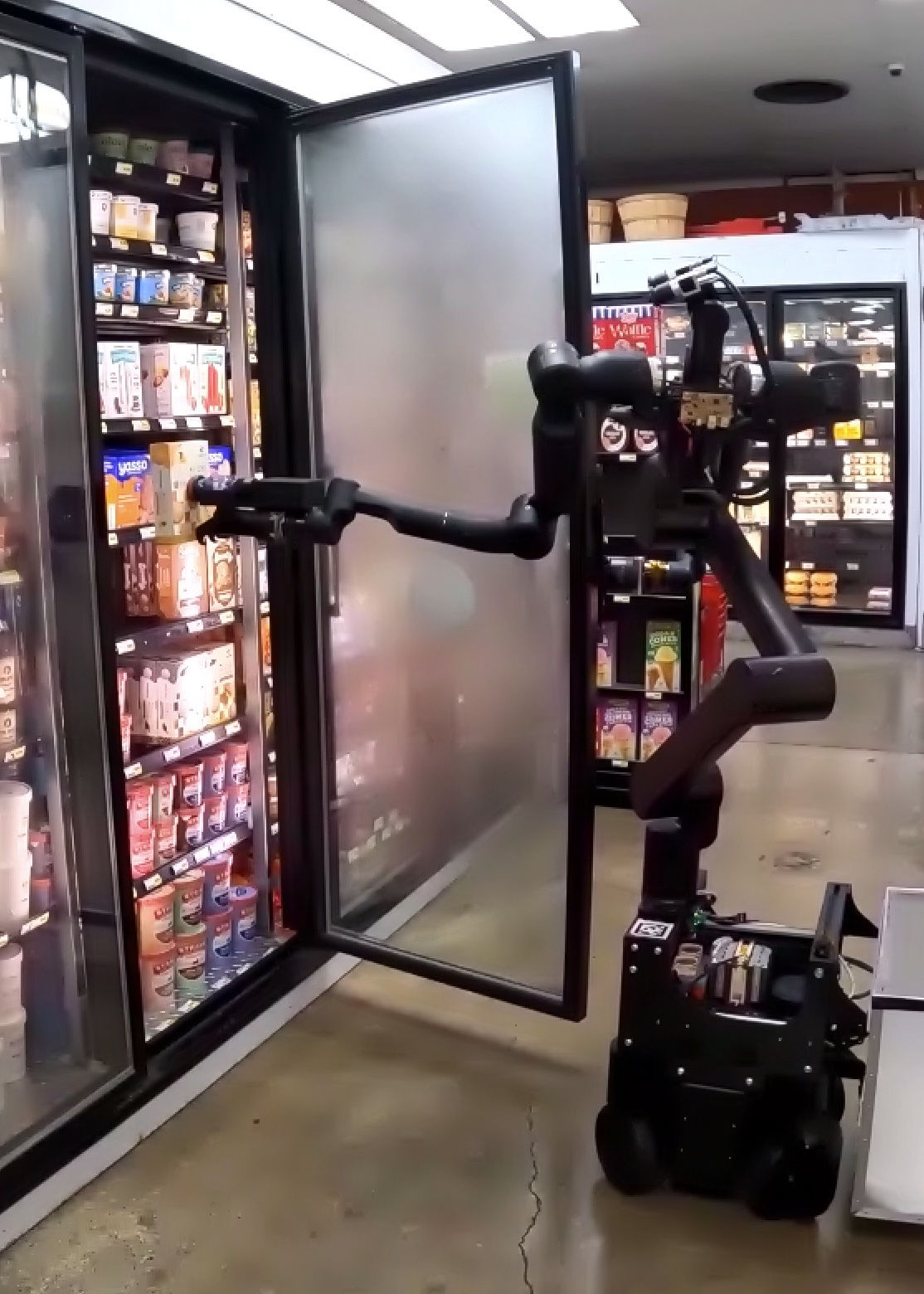}
        \caption{Fridges} 
        \label{fig:layout_b}
    \end{subfigure}%
    ~ 
    \begin{subfigure}{0.22\linewidth}
        \centering
        \includegraphics[width=20mm,scale=0.05, trim={0 0 0 0},clip]{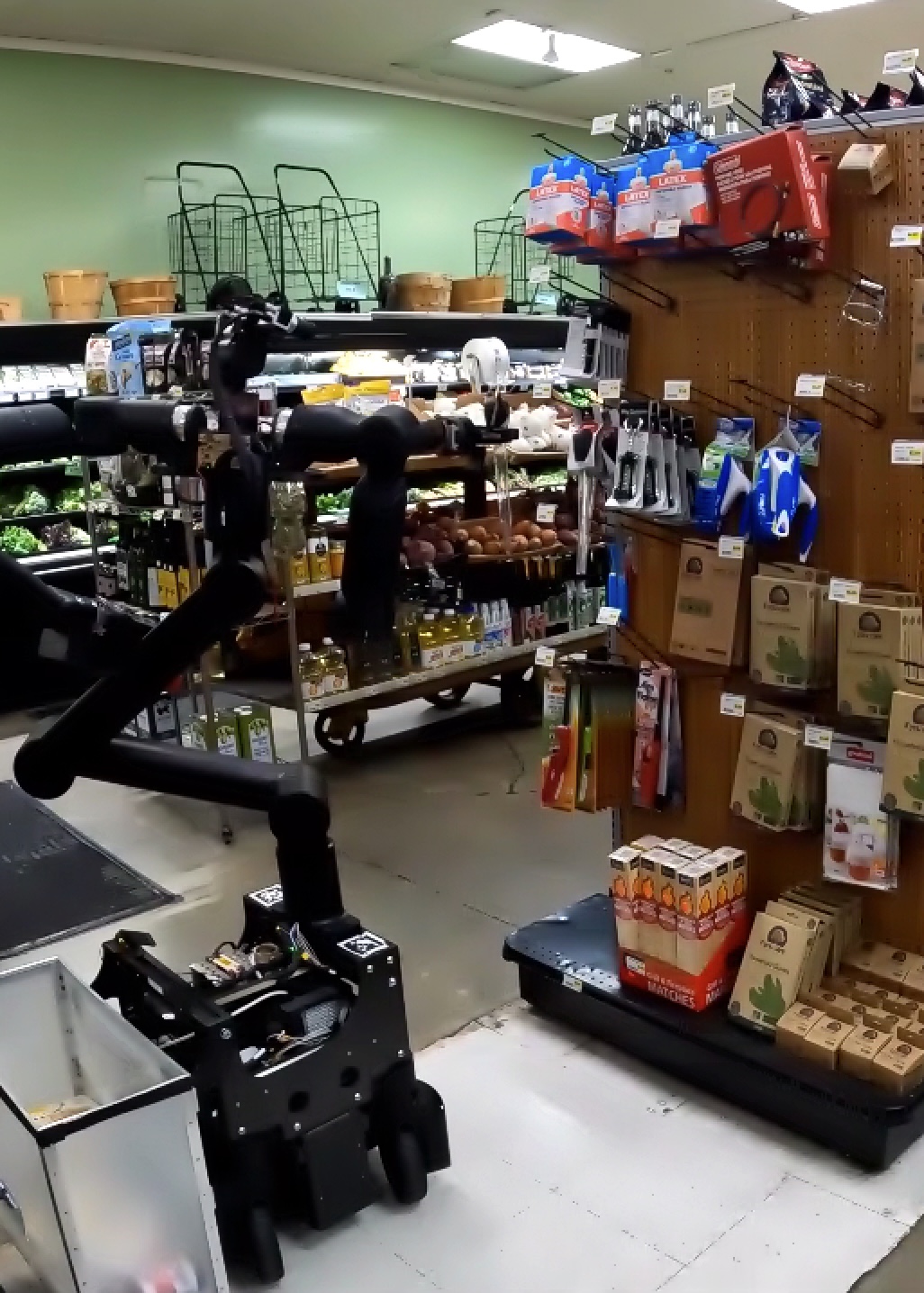}
        \caption{Hooks}
        \label{fig:layout_c}
    \end{subfigure}
    ~ 
    \begin{subfigure}{0.22\linewidth}
        \centering
        \includegraphics[width=20mm,scale=0.05, trim={0 0 0 0},clip]{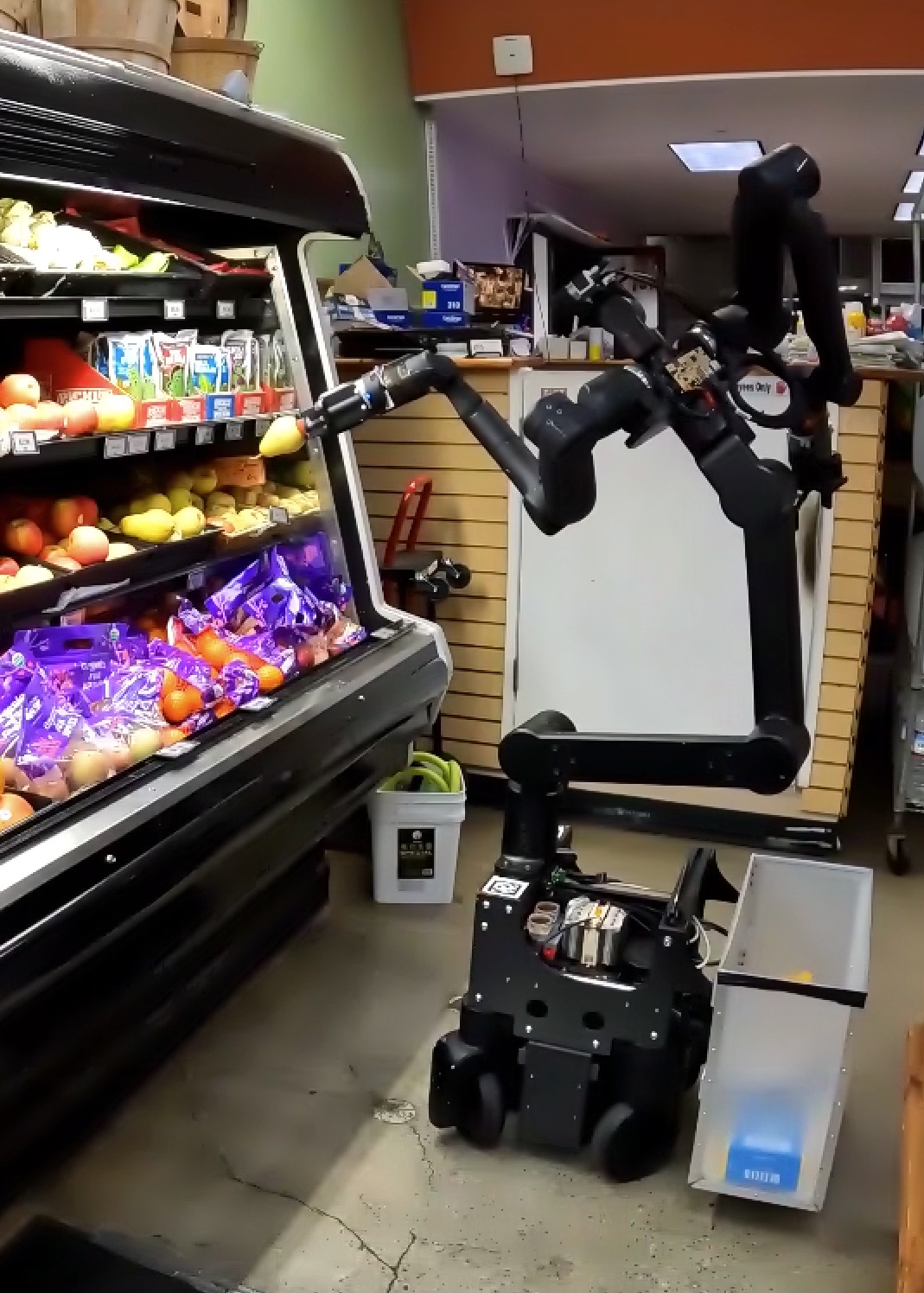}
        \caption{Piles}
        \label{fig:layout_d}
    \end{subfigure}
    \caption{SHOPPER retrieves items from a wide variety of different fixtures and item arrangements.}
    \label{fig:example_layouts}
\end{figure}

Our field tests are conducted in real grocery stores. A few days before the field test, a map of the environment is generated based on data collected with a mapping rig containing multiple stereo cameras. Using images from these stereo cameras, an offline map is generated with keyframes for online localization. An item detector and classifier are also run on the images, which allow us to obtain the location and the predicted UPC of every unique item and store it in the offline map. Further details on the mapping process can be found in \cite{MMT2023}. Of the 959 unique items automatically detected by our item detector, we exclude 236 from the graspable set because they are made of glass and pose a risk of permanent inventory damage. These items are arranged in a wide variety of layouts and fixtures, with some examples shown in Fig.~\ref{fig:example_layouts}.

\subsection{Online process}
We briefly outline the overall task flow. First, a path planner based on the Christofides algorithm generates a sequence of navigation targets that minimizes the path length required to visit all requested items. A navigation module using a variant of the Dynamic Window Approach (DWA) guides the robot between these points, handling dynamic obstacles and store layout changes (relative to the offline-generated map). At the start of the task, the robot is localized in the global map. Once localized, it uses visual odometry (VO) to update its pose while navigating to each item on the shopping list. To mitigate VO drift, the robot re-localizes with the global map each time it arrives at a new item. While the details of navigation and localization are outside the scope of this paper, we assume that these modules allow the robot to reach the approximate location of any requested item (within $\pm40$ cm).

\begin{figure}
\centering
\includegraphics[scale=0.5]{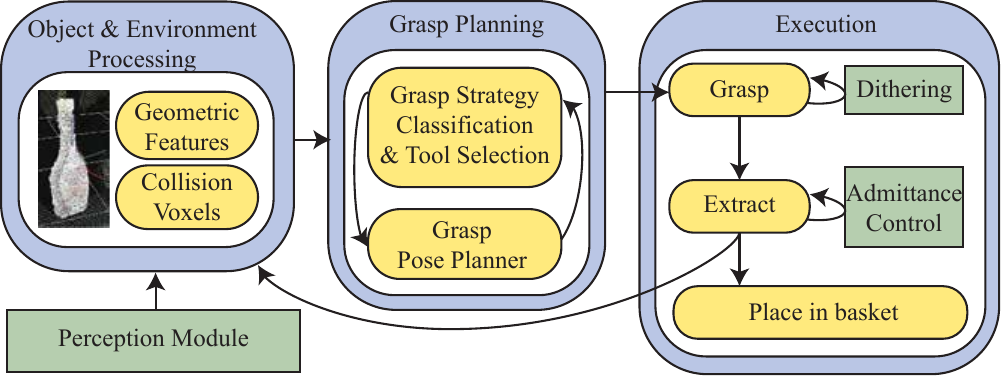}
\caption{Overview of the SHOPPER grasp module.}
\label{fig:grasp_module}
\end{figure}

\subsection{Grasp pipeline}
Once the robot has autonomously navigated in front of where it believes the item should be, it calls the perception module to locate the item. If the target item is not in sight, the robot simply moves to the next item in the shopping list. In addition to determining the existence of the target item, the perception module is also responsible for estimating the bounding box of the item, computing a segmented point cloud of the item, and also determining the pose of the item using Robust PCA\cite{NIPSNetrapalli2014} on a center patch of the item. 

The perception  module has three key components - a detector, a classifier, and a segmenter. The detector, based on DETR \cite{DETR}, is trained on fully synthetic data and outputs bounding boxes around grocery items in the images. 
The classifier then takes the bounding box output and matches the item to an item in an existing database or indicates that no match was found. The segmenter also takes the bounding box output and segments out the item. 

Figure~\ref{fig:grasp_module} illustrates what follows after the perception module.  From the segmented point cloud, relevant geometric features are extracted along with contextual information about the item's immediate surroundings, such as its fixturing type or potential collision voxels. The appropriate grasp strategy is then selected, which varies based on the chosen tool. These strategies, detailed in Section~\ref{sec:strategies}, dictate the approach to be taken. Next, a grasp pose planner generates the desired grasp pose along with pre-grasp waypoints. The robot then executes the grasp pose, attaching to the item either by closing the gripper or making contact with the item using the suction tool. Once attached, the robot extracts the item from the shelving infrastructure and places it in the basket behind it. If the grasp and extraction fail, the robot re-detects the item and restarts the entire perception and grasping process.

\section{STRATEGIES}\label{sec:strategies}

\begin{figure*}
\centering
\includegraphics[width=\textwidth]{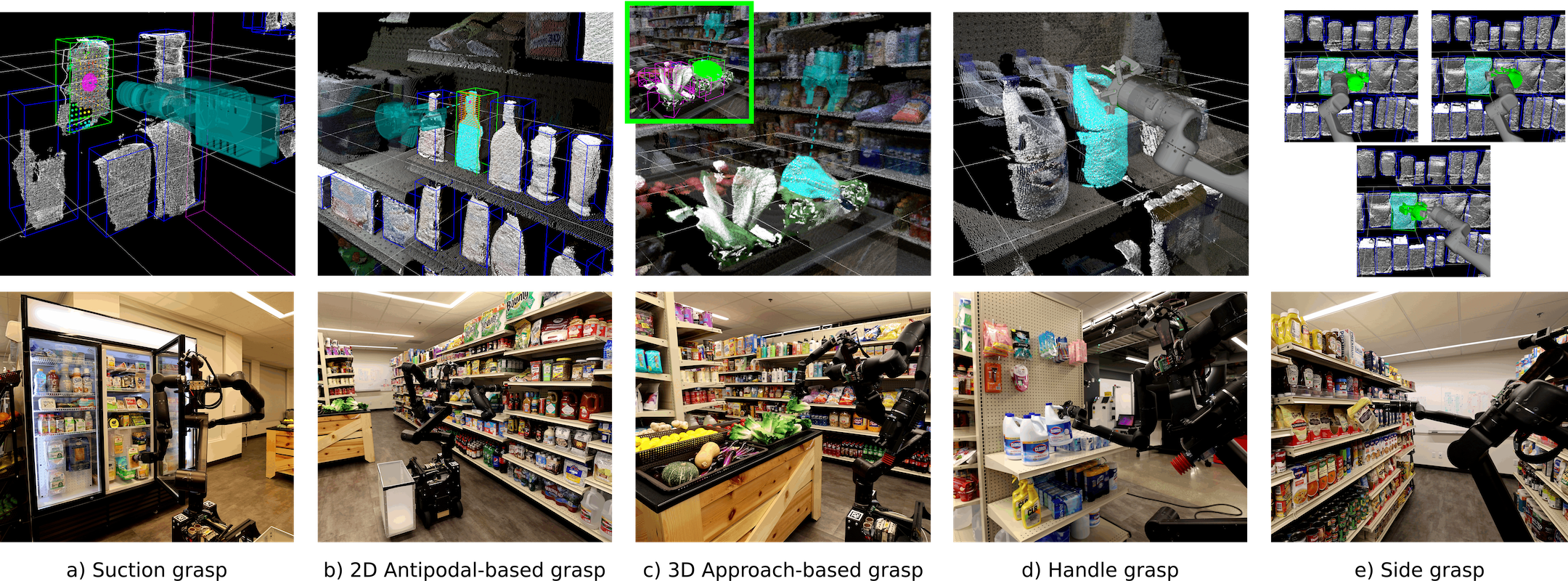}
\caption{Examples of the diverse range of grasp strategies with robot perception visualized at top: (a) a suction grasp that optimizes the most stable attachment point; (b) a gripper grasp that locates secure contact points using the 2D outline of the shelved item; (c) a gripper grasp planned with a shape-completed hull of the target item (see inset) for picking from produce piles; (d) a handled-item grasp that identifies keypoints to parameterize gripper insertion into the handle; (e) a side grasp designed for bags, where neighboring items are first swiped away to provide the gripper access to the bag's side.}
\label{fig:grasp_banner}
\end{figure*}

Grocery stores contain a vast assortment of items, ranging from small and lightweight products to heavy goods and bagged items. This diversity necessitates the use of \textit{generalizable} distinct grasping strategies, as we do not know the item shape and/or properties apriori nor do we have item models and must operate entirely off the output of our perception networks (e.g. detection, segmentation, classification). After substantial iteration, we found the most robust strategy was to categorize items into 6 broad categories, mapped to corresponding grasp strategies detailed below:

\noindent\textbf{2D Antipodal-based grasp.} For shelved items, we constrain the grasp approach to be perpendicular to the item's front face. We then project each segmented item point cloud onto this plane and compute a 2D concave hull around the shape. Antipodal grasp points are sampled randomly along the hull's boundary~\cite{EppnerISRR2019}. The candidate point pair that falls within the gripper's stroke width and is farthest from potential collisions is then selected as the optimal grasp. An example grasp is in Fig.~\ref{fig:grasp_banner}b.

\noindent\textbf{Approach-based grasp}. For unshelved produce items, we cannot constrain the grasp approach pose, and the grasping problem shifts from 2D to 3D. Therefore, a 3D ellipsoidal hull is extrapolated using the partially-observed point cloud of the target object, and grasp sampling is done in three dimensions around this hull. See Fig.~\ref{fig:grasp_banner}c for an example. % While antipodal grasps are typically more stable due to near-perpendicular contact normals, they are highly prone to collisions when dealing with piled items. As a result, very few antipodal grasps are viable in produce-grasping scenarios, and we choose to use approach-based sampling instead~\cite{EppnerISRR2019}.

\noindent\textbf{Handled-Item grasp}. For heavy objects with handles, grasping at the handle provides the most stable hold by leveraging finger support. To accomplish this, we fuse a handle keypoint detector with geometric grasp sampling about a 2D concave hull fit to the handle opening. (if no hole is seen, the keypoint determines the grasp location). An example of this grasp is in Fig.~\ref{fig:grasp_banner}d. %Grasps are planned such that one finger is inside the opening and the other is outside the handle for stability. For such items, some of which can exceed 3 kg, we impose a strict tool pose constraint to prevent slippage throughout the pick-and-place process.

\noindent\textbf{Pinch grasp}. Some items, like short cans, are too small for suction contact and too wide for the 2D gripper. In these cases, we pinch the top front edge of the item between the gripper fingers.

\noindent\textbf{Side grasp}. Many items in grocery stores are too wide for the 2D gripper planner and incompatible with the suction tool (e.g. material too porous or shear forces too large). In these cases, we grasp the item from the side, using intermediate waypoints to push aside any neighboring items that block access to the target (Fig.~\ref{fig:grasp_banner}e). The side to grasp (i.e., left or right) is chosen to minimize collisions with nearby items or fixtures. % The bag materials are often porous, too soft to form a good seal, or too heavy, causing shear forces that detach the item from the suction cup immediately. In these cases, we grasp the bags from the side, using intermediate waypoints to push aside any neighboring items that block access to the target (Fig.~\ref{fig:grasp_banner}e). The side to grasp (i.e., left or right) is chosen to minimize collisions with nearby items or fixtures. Although the ideal approach angle is perpendicular to the item face, it is adjusted during runtime to avoid interference with surrounding obstacles.

\noindent\textbf{Suction grasp.} To determine optimal suction points, our algorithm computes a 2D concave hull around the item. Candidate suction points are sampled on the interior, and are ranked based on (1) signed distance from the edge, and (2) flatness in the local region. The highest-ranked collision-free candidate with sufficient flatness is selected. An example of a suction grasp is in Fig.~\ref{fig:grasp_banner}a. %The projection plane depends on context -- vertical for shelved or hanging items and parallel to the camera for loose produce. The concave hull defines signed distances from its edge, which are used to rank suction locations. For shelved items, the ranking favors greater signed distances while biasing toward the top of the item to minimize moment loads under angular deflection at the suction interface, a crucial factor for heavier items. For hanging objects or piled produce, the ranking is based solely on signed distance. Suction points are then evaluated consecutively, using a flatness metric for grasp quality. The first collision-free candidate with sufficient flatness as compared to prior candidates is selected. An example of a suction grasp is in Fig.~\ref{fig:grasp_banner}a.

~

Surprisingly, these 6 strategies cover the majority of items we see in the store. Determining which one to use is done online through a combination of learning (e.g. neural networks to classify the object type) and geometric reasoning (considering the item dimensions). We found this to be more robust than a learning-only approach.

\section{CHALLENGES}

In this section, we highlight key grasping-related challenges encountered during our development process. For each challenge, we outline the strategies and algorithms used to address and mitigate these issues.

\subsection{Imprecision}

Mobile manipulation is invariably affected by imprecision. In contrast to industrial arms bolted down, mobile robots must deal with kinematic errors in their arms and torso (in addition to ever-present sensor noise). Because each kinematic chain in our robot contains 12 joints from the base of the torso to the gripper, build up of error accumulates, ultimately reducing the likelihood of a successful grasp. To address these issues, our approach incorporates several mitigation strategies.

\noindent\textbf{Dithering}.
If the suction grasp is missed during the initial approach due to kinematic imprecision, the robot executes small, probing movements—referred to as dithering—around the initial target pose. This process continues until the suction tool detects a pressure signal indicating a successful grasp or until a preset timeout occurs. To ensure that the tool does not exert excessive force on the environment, we employ admittance control \cite{maples1986experiments, newman1992stability}. Notably, during the latest field test, 14 out of 17 instances in which the suction grasp initially failed were successfully recovered through dithering. %The motion involves moving the tool back and forth while slightly varying its position in a direction perpendicular to the primary motion. To ensure that the tool does not exert excessive force on the environment, we employ admittance control \cite{maples1986experiments, newman1992stability}, which models the tool’s motion as a virtual mass-spring-damper system, enabling compliance against external forces. Notably, during the latest field test, 14 out of 17 instances in which the suction grasp initially failed were successfully recovered through dithering.

\noindent\textbf{Tool pose correction}. Good hardware/calibration can often reduce error in gripper pose to centimeter-level accuracy. However, in many cases, sub-centimeter accuracy is required for success. To accomplish this, we add a kinematic correction \textit{at runtime} by comparing the segmented point cloud for our gripper to a CAD model of our gripper. We run point cloud registration to compute a pose delta between the observed point cloud and a reference point cloud (produced by sampling the CAD model of our gripper). The details of the point cloud registration algorithm, DSES, can be found in \cite{DSES2024}. This pipeline helps ensure that the grasp pose we command the robot to is accurately achieved. % However, it should be noted that tool pose correction cannot remedy kinematic errors in the torso.

\noindent\textbf{ICP alignment of camera sources}. Due to imperfect kinematic calibration or disturbances caused by accidental collisions, the point clouds generated by the head and chassis cameras may not align perfectly by default. However, alignment is crucial when the grasp depends on perception from multiple sources. In these cases, an Iterative Closest Point (ICP) algorithm is used to align the point clouds from both sources, alleviating any misalignment. %(e.g., item detection from the chassis cameras but tool pose correction from the head cameras). In these cases, an Iterative Closest Point (ICP) algorithm is used to align the point clouds from both sources, alleviating any misalignment.

\subsection{Environmental Obstacles} \label{sec:env_obs}

\begin{figure}
\centering
\includegraphics[width=0.8\linewidth]{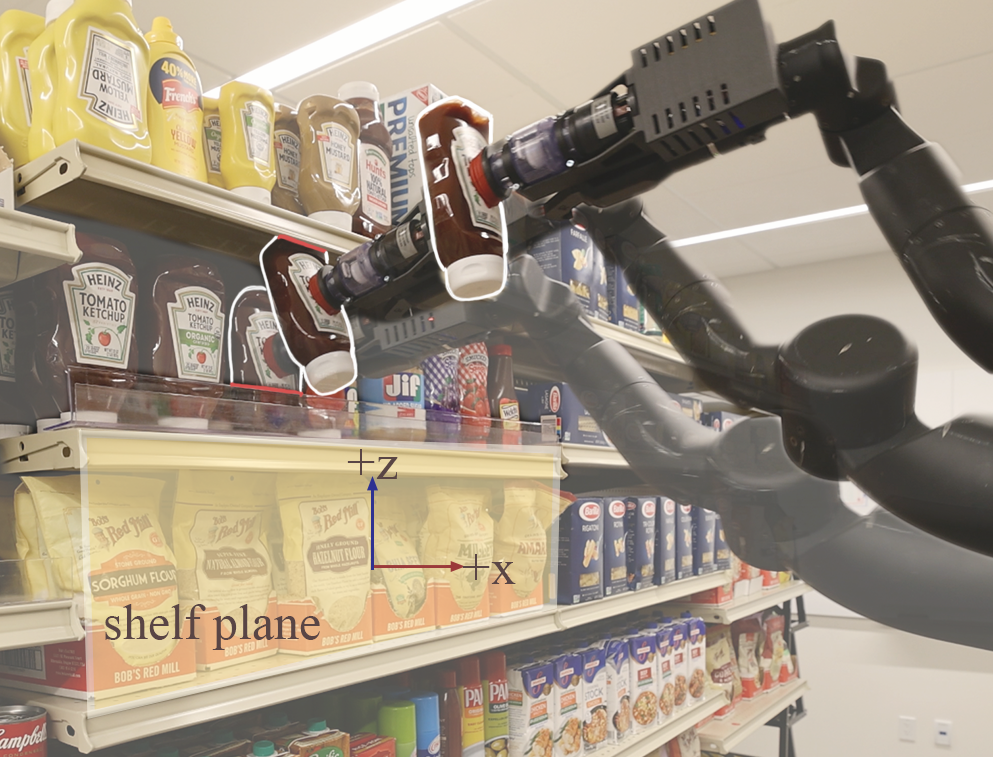} 
\caption{Extraction of a shelf item behind a clear lip.}
\label{fig:shelf_lip}
\end{figure}

A significant challenge in our task is operating in tight, complex collision environments, dealing with elements such as shelf lips, grates, and hooks, as depicted in Fig.~\ref{fig:shelf_lip}. These structures—whether inherent to the shelving units or part of the objects themselves—create additional obstacles in already tight spaces that complicate the grasping process. These obstacles are particularly problematic during the extraction phase. As the robot extracts an item from the shelf, there is a risk that the grasped object may collide with these elements, leading to a loss of grip or unintended shifts of the item, ultimately causing extraction failure. 

\noindent\textbf{Custom admittance control}. To mitigate these challenges, our system employs an admittance control strategy specifically tuned for the extraction phase. This control method ensures compliant interaction with the environment and leverages force feedback to actively adjust the tool’s trajectory away from obstacles when contact occurs. We achieve this by introducing a non-diagonal term in the stiffness matrix of the admittance controller. Typically, the stiffness matrix is diagonal, meaning that position deviations from the target in one direction only affect the motion in that same direction. In contrast, our design includes a non-diagonal coupling term that effectively translates a deviation in one direction into motion in another. In particular, when the extraction motion is impeded and a positional deviation is detected along the extraction direction—$\mathbf{x}$-axis in Fig.~\ref{fig:shelf_lip}, the non-diagonal term generates acceleration in the upper direction—$\mathbf{z}$-axis in Fig.~\ref{fig:shelf_lip}—moving the tool away from the obstacles. This cross-coupling enables the robot to adapt its trajectory, moving upward to navigate over the obstacle. % while maintaining a secure grasp on the item.

\noindent\textbf{Opening doors}. Many grocery items are inside of refrigerators and freezers, which requires the robot to have door opening capabilities in order to access them. By itself this poses a manipulation challenge just to open the door. However, afterwards, the door opening arm is constrained to hold the door open during the pick execution. Additionally, the fridge doors that we manipulate are typically weighted to automatically close unless held open, so continuous contact must be preserved during the entire manipulation phase.

For opening the door we mount a custom hook onto our tools and use a contact-based heuristic to guide the hook into the handle. Then, we use the known hinge axis relative to the handle to generate a set of waypoints along the expected door opening arc for the desired position path and then use admittance control in the radial axis to help ensure a steady hold on the handle. While planning the target pick, we still use a joint-level motion planner, but during execution we employ a parallel reactive Cartesian admittance controller on the door-manipulating arm that holds the arm tip still while pulling into the handle to hold the door open in place.

As an additional consequence of having one arm constrained in a parallel task, we must remove that tool's grasp strategies from the set of available grasps.

\subsection{Scene understanding} To avoid collisions, it is crucial to identify collision voxels within a scene. At the same time, hard obstacles (e.g., shelving or other infrastructure) and soft obstacles (e.g., neighboring items that can be manipulated or pushed aside) must also be differentiated. Using our item processing module to segment all grocery items, we classify non-target item voxels as soft obstacles, while all other voxels are considered hard obstacles.

\subsection{Partial observations} 
Due to occlusions from shelving or adjacent items, SHOPPER only has access to partial observability of the target item. While many grasps can be planned using solely the information from the item's partial point cloud, shape completion can be critical for reasoning over unobserved parts of the item. To plan 3D gripper grasps over produce items, we make the assumption that produce items are approximately ellipsoid, and calculate the minimum volume ellipsoid that contains all points. However, a reliable, general shape completion strategy could alleviate this challenge and improve robustness in these instances. %~\cite{ellipsoid}.

\section{ANALYSIS \& INSIGHTS}
We conduct our field test experiments in a real, unmodified grocery store four times per year. However, in this work we discuss only the results from the latest field test in order to report only on the most up-to-date results and item scope. Throughout the years, our system and task have constantly been improving, not only in the sophistication of our hardware and algorithms, but also in the breadth of items we allow the SHOPPER robot to attempt. For discussion on the evolution of our challenge task and capabilities over time, please see our website\footnote{\url{https://tri-ml.github.io/shopper}}. 

In this section, we first present a detailed breakdown of the failure modes experienced during this last field test. Then, we summarize our key takeaways and highlight what we believe the most fundamental challenges are. By sharing these practical insights, we aim to help the robotics community identify major limitations hindering general-purpose mobile manipulation. We hope that these insights will guide the direction of future development. % whether using traditional methods, end-to-end learning, or something in between.

\subsection{Failure Analysis}
During our latest field test, we collected data over three nights, totaling more than 10 hours of robot runtime. During this period, SHOPPER attempted 335 distinct pick attempts, with a 61\% success rate, meaning the target item was successfully placed in the robot’s shopping basket. We identify eight fundamental root causes ($R1$–$R8$) behind all pick failures observed in our experiments. Table~\ref{tab:tool_failure_breakdown} illustrates the overall distribution of these failure causes, along with their frequency for both the suction and gripper tools separately.

\newcommand{\firstcolor}{\cellcolor[HTML]{e37680}$R1$}
\newcommand{\secondcolor}{\cellcolor[HTML]{cd233e}$R2$}
\newcommand{\thirdcolor}{\cellcolor[HTML]{e76f51}$R3$}
\newcommand{\fourthcolor}{\cellcolor[HTML]{f4a261}$R4$}
\newcommand{\fifthcolor}{\cellcolor[HTML]{e9c46a}$R5$}
\newcommand{\sixthcolor}{\cellcolor[HTML]{2a9d8f}$R6$}
\newcommand{\seventhcolor}{\cellcolor[HTML]{264653}$R7$}
\newcommand{\eigthcolor}{\cellcolor[HTML]{875984}$R8$}

\begin{table*}[!ht]
    \centering
    \begin{tabular}{p{3cm}p{0.32cm}p{7cm}p{1.2cm}p{1.4cm}p{1.4cm}}
    \toprule
        ~ & ~ & Failure mode & Total (\%) & Suction (\%) & Gripper (\%)\\ \midrule
        \multirow{8}{*}{ \includegraphics[scale=0.2]{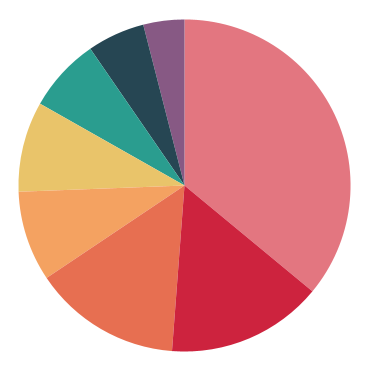}} & \firstcolor & Heavy, bulky, and incompatible items & \multicolumn{1}{r}{36.0} & \multicolumn{1}{r}{45.6} & \multicolumn{1}{r}{19.6} \\ 
        & \secondcolor & Edge cases in motion and grasp planning algorithms & \multicolumn{1}{r}{15.2} & \multicolumn{1}{r}{22.8} & \multicolumn{1}{r}{2.2} \\ %\hline
        & \thirdcolor & Unexpectedly displaced target item  & \multicolumn{1}{r}{14.4} & \multicolumn{1}{r}{\textendash} & \multicolumn{1}{r}{39.1}\\ %\hline
        & \fourthcolor & Difficult extract & \multicolumn{1}{r}{8.8} & \multicolumn{1}{r}{11.4} & \multicolumn{1}{r}{4.3} \\ %\hline
        & \fifthcolor & Interference from neighboring item & \multicolumn{1}{r}{8.8} & \multicolumn{1}{r}{\textendash} & \multicolumn{1}{r}{23.9} \\ %\hline
        & \sixthcolor & Fell on the way to basket  & \multicolumn{1}{r}{7.2} & \multicolumn{1}{r}{11.4} & \multicolumn{1}{r}{\textendash} \\ 
        & \seventhcolor & Perception (missed collision voxels, poor item segmentation) & \multicolumn{1}{r}{5.6} & \multicolumn{1}{r}{6.3} & \multicolumn{1}{r}{4.3} \\ 
        & \eigthcolor & Kinematic imprecision & \multicolumn{1}{r}{4.0} & \multicolumn{1}{r}{2.5} & \multicolumn{1}{r}{6.5} \\ \bottomrule
    \end{tabular}
    \caption{Breakdown of overall pick failure modes across all grasps ($R1$–$R8$), along with separate failure breakdowns for each tool type (i.e., suction and gripper).}
    \label{tab:tool_failure_breakdown}
\end{table*}

\begin{figure}
\centering
\includegraphics[scale=0.51]{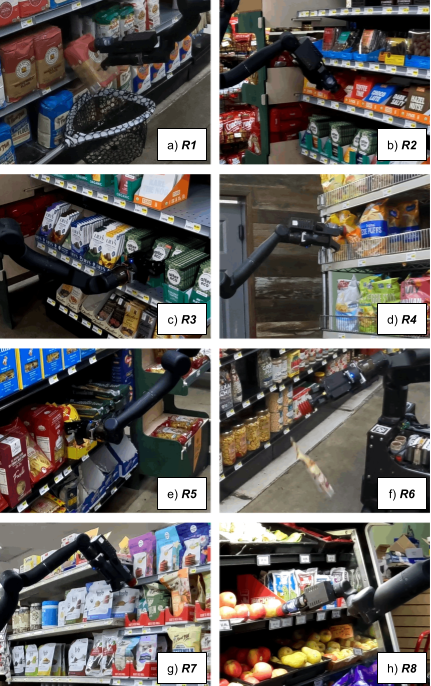}
\caption{Examples of each failure mode listed in Table~\ref{tab:tool_failure_breakdown}: (a) a heavy flour bag that fails with all available grasp strategies; (b) a grasp pose-planning edge case where the only collision-free grasp fails to create a stable seal on a discontinuous surface; (c) a target item is displaced from the gripper during grasp execution, making it ungraspable at the planned pose; (d) a challenging extraction case where the item is taller than the available opening; (e) a failed side grasp due to the gripper being unable to swipe away a tightly packed adjacent item; (f) a grasp that remains stable until dynamic motions during placement into the basket cause failure; (g) a collision with a shelf due to undetected collision voxels; and (h) imprecise kinematic execution resulting in failure to achieve the planned grasp pose.}
\label{fig:fail_examples}
\end{figure}

\subsection{Limitations \& Insights}
Our failure analysis highlights fundamental limitations that prevent SHOPPER from achieving true general-purpose mobile manipulation. We discuss these challenges here and identify the most promising areas for future development.

\noindent\textbf{Heavy or bulky items}. Many grocery store items remain incompatible with our current grasp strategies, accounting for 36\% of failed picks ($R1$). SHOPPER struggles with items that are either too heavy for both the suction tool and gripper or too bulky for a stable grasp, such as dense flour bags, heavy soda boxes, and large paper towel packs (Fig.~\ref{fig:fail_examples}a). Addressing these failures requires pre-contact weight estimation and dual-arm coordination, as even humans often need both hands to grasp heavy and bulky items. However, dual-arm picking introduces significant challenges: motion planning becomes exponentially more complex, and once an item is grasped, the arms form a closed kinematic chain, further constraining movement. Additionally, precise force control is crucial to maintain a secure grip while preventing excessive internal forces that could destabilize the system. To tackle these challenges, strategies such as grasping objects with both arms on either side \cite{wu2025wild} and pulling heavy objects onto a supporting arm have been explored. While promising, these approaches require further refinement before full integration into SHOPPER’s task pipeline.

\noindent\textbf{Incompatible items}. Beyond heavy or bulky items, there are several edge-case items that are difficult to grasp. Articulated produce frequently fail due to their fragile structure—parts may break off (e.g., a cabbage leaf) or SHOPPER may fail to recognize that the target is part of a larger cluster (e.g., one banana in a bunch). Currently, SHOPPER does not infer object articulation properties, preventing it from anticipating how such items will respond to a planned grasp. Improving this capability would unlock more advanced manipulation.

\noindent\textbf{Lack of real-time item state tracking.} A significant portion of all failure modes stems from the fact that SHOPPER's grasp execution is nearly entirely open-loop. As a result, if an item is unexpectedly displaced during grasp execution ($R3$, e.g., Fig.~\ref{fig:fail_examples}c), there are no corrective interventions since SHOPPER cannot detect the imminent failure. These interventions could be enabled through real-time item state tracking. However, adding to the challenge, oftentimes the item becomes heavily (or fully) obscured from the cameras during grasping. Adding more cameras or tactile sensors could be crucial to support continuous tracking. Tracking neighboring items is also crucial, particularly when interaction with them is expected (e.g., during side grasps). For example, SHOPPER may attempt to swipe a neighboring item away but fail due to its weight or tight packing, leaving the neighboring item still blocking access to the target and causing an unexpected failure ($R5$, e.g., Fig.~\ref{fig:fail_examples}e). Another failure mode occurs when an item is securely extracted from the shelf but falls off the tool on its way to the basket due to dynamic instability ($R6$, e.g., Fig.~\ref{fig:fail_examples}f). These failures go unnoticed due to the lack of item state tracking during the final movement toward the basket.

\noindent\textbf{Difficult extract scenarios}. Although we employ admittance control to mitigate environmental obstacles (Section~\ref{sec:env_obs}), certain shelf configurations remain problematic ($R4$). In some cases, the gap between the upper and lower lips of the shelves is so narrow—often even less than the height of the item, (e.g., Fig.~\ref{fig:fail_examples}d)—that the items get stuck between and result in failed extractions. To address this, we plan to incorporate enhanced rotational compliance during the extraction phase. By allowing the item to tilt as it is being extracted, the robot can better align the item with the available clearance.

\noindent\textbf{Perception and kinematic noise.} Designing robust strategies in the presence of imperfect modules is a fundamental challenge. Occasionally, perception noise ($R7$) causes artifacts such as missing voxels. Missing item voxels can lead to inaccurate item geometry inference, while missing shelf voxels may result in unexpected collisions (e.g., Fig.\ref{fig:fail_examples}g). Kinematic imprecision can cause the actual grasp pose to differ from the planned one ($R8$). In more challenging cases, where tool pose correction is not possible due to a completely blocked view of the arm, this can result in grasp attempts that miss the item entirely (e.g., Fig.\ref{fig:fail_examples}h).

\noindent\textbf{Edge cases in planning algorithms} Real-world environments pose challenges due to spatial constraints, which can violate assumptions made in our grasp planning algorithms or create kinematically complex scenarios that strain the planner ($R2$). For example, we assume that the optimal suction grasp makes perpendicular contact with the item's surface. However, no collision-free grasps are possible under such a requirement for the example shown in Fig.~\ref{fig:fail_examples}b), where a chocolate bar is tilted with its front surface facing upwards. Furthermore, items positioned behind large fixtures (e.g., a fridge's center divider) or stocked on hard-to-reach shelves, such as low shelves in narrow aisles, can severely challenge motion planning. In such cases, pre-manipulation or improved base planning may be necessary to position the item relative to the robot in a more manipulable configuration. 
\section{DATASET}
We further provide a dataset from three unique field testing events that span two real-world unseen grocery stores\footnote{\url{https://tri-ml.github.io/shopper}}. This dataset covers 1200+ pick attempts, labeled as success or failure, over 800+ unique items. The grasp data includes robot base and arm actions as well as 3D perception from both the head and chassis stereo cameras. We also additionally provide navigation data between items in the grocery stores. We hope that this dataset can provide helpful data for further analysis, or training data (for pre-training of robot foundational models, for learning from failure data, etc.).

\section{DISCUSSION}
In this work we present SHOPPER, a general-purpose mobile manipulation robot, and detail the grasp strategies we have implemented to enhance its robustness and generalizability in real-world scenarios. Through an in-depth analysis of our latest large-scale grocery shopping field test—spanning over 10 hours of robot operation— we identify key limitations preventing us from achieving 100\% pick success. This ambitious benchmark highlights both overlooked edge cases and fundamental capabilities yet to be robustly implemented, such as reliable object state tracking and reliable dual-arm coordination. We believe SHOPPER remains an invaluable test platform for advancing general manipulation strategies, which extend beyond grocery shopping to other complex tasks like item restocking. Moreover, beyond serving as a cutting-edge platform for algorithm development, SHOPPER’s field test data presents an exciting opportunity for foundation model training. By autonomously generating demonstrations without relying on human teleoperation, SHOPPER enables high-quality, scalable data collection for robot learning.

%%%%%%%%%%%%%%%%%%%%%%%%%%%%%%%%%%%%%%%%%%%%%%%%%%%%%%%%%%%%%%%%%%%%%%%%%%%%%%%%

\addtolength{\textheight}{-12cm}   % This command serves to balance the column lengths

%%%%%%%%%%%%%%%%%%%%%%%%%%%%%%%%%%%%%%%%%%%%%%%%%%%%%%%%%%%%%%%%%%%%%%%%%%%%%%%%

%%%%%%%%%%%%%%%%%%%%%%%%%%%%%%%%%%%%%%%%%%%%%%%%%%%%%%%%%%%%%%%%%%%%%%%%%%%%%%%%

%%%%%%%%%%%%%%%%%%%%%%%%%%%%%%%%%%%%%%%%%%%%%%%%%%%%%%%%%%%%%%%%%%%%%%%%%%%%%%%%

\printbibliography

\end{document}